%% file: root.tex
\documentclass[letterpaper, 10 pt, conference]{ieeeconf}  
\IEEEoverridecommandlockouts                              

\usepackage[english]{babel}
\usepackage{csquotes}

\usepackage[style=ieee,doi=false,isbn=false,url=false,eprint=false]{biblatex}

\AtBeginBibliography{\small}

\usepackage{amsmath}
\usepackage{amssymb}
\usepackage{tabularx}
\usepackage{array}
\usepackage{graphicx}
\usepackage[hidelinks]{hyperref}
\usepackage{cleveref}

\usepackage[dvipsnames]{xcolor}
\usepackage{booktabs}
\usepackage{placeins}
\usepackage[scr=boondox,  
            cal=esstix]   
           {}

\usepackage{algorithm}
\usepackage{algorithmic}

\usepackage{booktabs}
\usepackage{siunitx} 
\usepackage{bm}       

\usepackage[compact]{titlesec}
\titlespacing*{\section}{0pt}{*0.8}{*0.8}
\titlespacing*{\subsection}{0pt}{*0.7}{*0.7}
\usepackage[font=small,skip=5pt]{caption}

\title{\LARGE \bf
Sampling-Based Model Predictive Control for Dexterous \\ Manipulation on a Biomimetic Tendon-Driven Hand
}

\author{Adrian Hess$^{1}$, Alexander M. Kübler$^{1}$, Benedek Forrai$^{1}$, Mehmet Dogar$^{2*}$ and Robert K. Katzschmann$^{1*}$
\thanks{$^{1}$Soft Robotics Lab, Department of Mechanical and Process Engineering, ETH Zurich, Switzerland}%
\thanks{$^{2}$School of Computer Science, University of Leeds, UK}%
\thanks{$^{*}$Emails: {\tt\small\href{mailto:M.R.Dogar@leeds.ac.uk}{m.r.dogar@leeds.ac.uk}} and {\tt\small \href{mailto:rkk@ethz.ch}{rkk@ethz.ch}} }%
}

\begin{document}

\maketitle
\thispagestyle{empty}
\pagestyle{empty}

\begin{abstract}
\input{Content/01-abstract}
\end{abstract}

\input{Content/02-introduction}
\input{Content/03-methodology}
\input{Content/04-experiments-results}
\input{Content/05-conclusion}

\section{Acknowledgments}
We greatly thank Chenyu Yang for his support in the project. This work was funded by an NCCR Startup Grant, an ETH Zurich RobotX Research Grant, a Hasler Foundation Grant, and Credit Suisse. M. Dogar was supported by EPSRC (EP/V052659/1).


\addtolength{\textheight}{0cm}   



\FloatBarrier




\printbibliography[title={References}]


\end{document}

%% file: Content/01-abstract.tex
Biomimetic and compliant robotic hands offer the potential for human-like dexterity, but controlling them is challenging due to high dimensionality, complex contact interactions, and uncertainties in state estimation. Sampling-based model predictive control (MPC), using a physics simulator as the dynamics model, is a promising approach for generating contact-rich behavior. However, sampling-based MPC has yet to be evaluated on physical (non-simulated) robotic hands, particularly on compliant hands with state uncertainties. We present the first successful demonstration of in-hand manipulation on a physical biomimetic tendon-driven robot hand using sampling-based MPC. While sampling-based MPC does not require lengthy training cycles like reinforcement learning approaches, it still necessitates adapting the task-specific objective function to ensure robust behavior execution on physical hardware. To adapt the objective function, we integrate a visual language model (VLM) with a real-time optimizer (MuJoCo MPC). We provide the VLM with a high-level human language description of the task and a video of the hand's current behavior. The VLM gradually adapts the objective function, allowing for efficient behavior generation, with each iteration taking less than two minutes. We show the feasibility of ball rolling, flipping, and catching using both simulated and physical robot hands. Our results demonstrate that sampling-based MPC is a promising approach for generating dexterous manipulation skills on biomimetic hands without extensive training cycles.\footnote{Video with experiments: \url{https://youtu.be/u4d6v3ohsOI}}

%% file: Content/02-introduction.tex
\section{Introduction}

Robotic hands have the potential to achieve human-like dexterity, but effectively controlling them poses significant challenges due to their high-dimensionality and complex contact states. Biomimetic hands with compliant actuation offer advantages; however, they also bring additional challenges, \textit{e.g.}, the difficulty of estimating the state of compliant links. Overcoming these challenges would allow biomimetic robotic hands to become the ideal universal manipulation platform in human-centric environments. Current reinforcement learning (RL) methods have shown success in dexterous manipulation tasks on robotic hands, but are limited by long training times and the need for task-specific retraining. Imitation learning (IL) is also emerging as an alternative, but as of now, it still requires abundant robot teleoperation data to master dexterous manipulation tasks.

\begin{figure}[ht]
   \centering
   \includegraphics[width=1\columnwidth]{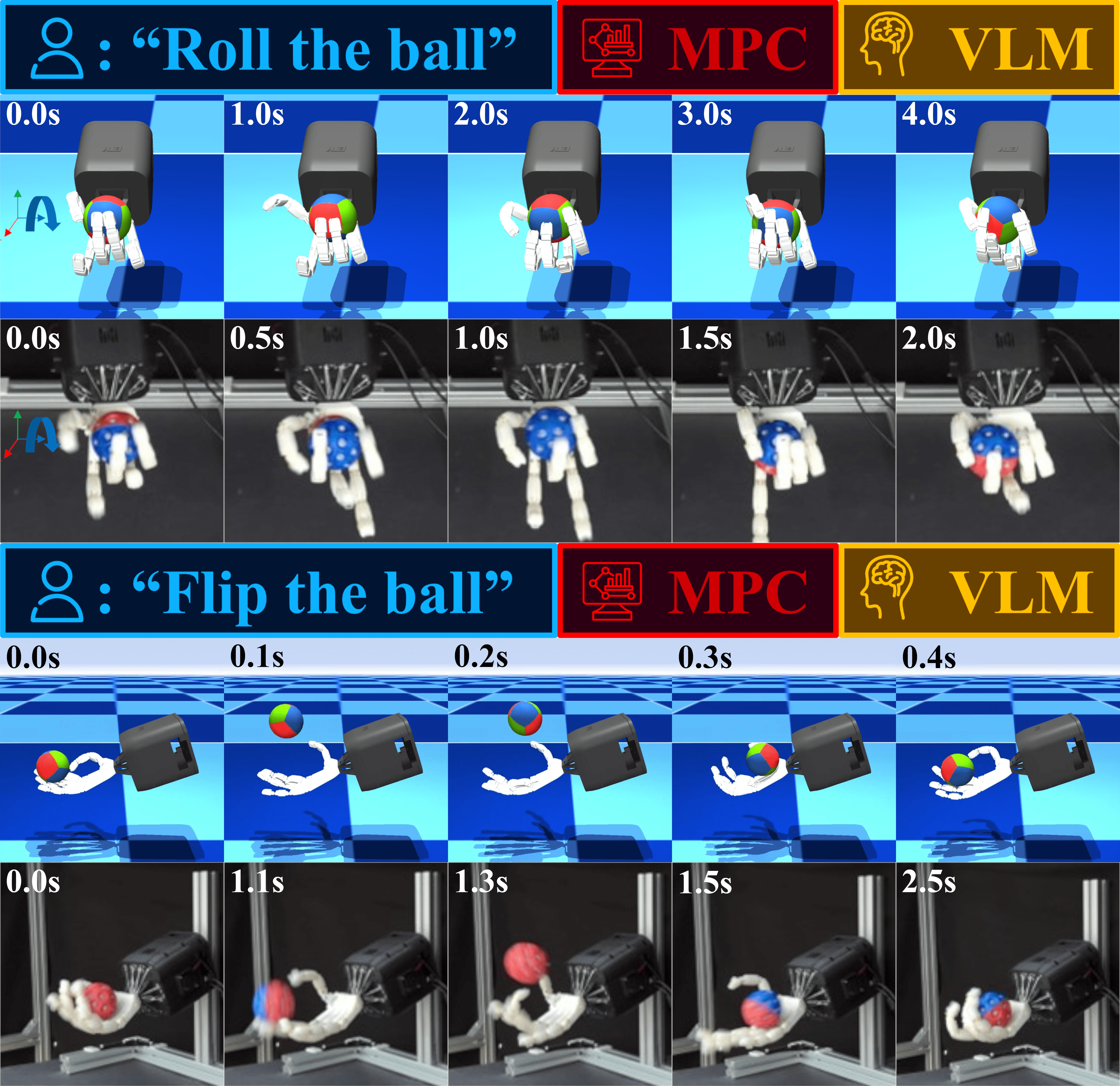}
   \caption{Our system accepts a human language description of the task, which is used by a VLM to adapt the objective function of a model predictive controller. We show demonstrations of in-hand ball rolling and ball flipping, both in simulation and on the physical robot hand. Different timestamps are used to display the results in the simulated and real environments.}
   \label{fig:overview}
\end{figure}

Sampling-based MPC offers a flexible alternative that excels in contact-rich manipulation tasks without the need for retraining. However, its application has been largely limited to simulations and has not yet been shown to work on physical robot hands. Our approach combines the ability of sampling-based MPC to execute new tasks without requiring retraining with the flexibility of a VLM. The VLM translates simple language commands into task-specific cost functions for the MPC, enabling immediate task execution, visual performance assessment, and rapid autonomous iteration for the continuous refinement of manipulation behaviors. Integrating these components closes the gap between simulation and real-world deployment, facilitating robust and adaptive manipulation with a biomimetic, tendon-driven robotic hand (see \Cref{fig:overview} for an overview).

\subsection{State of the art in dexterous manipulation}

Learning-based methods have shown great promise in dexterous manipulation. OpenAI~\cite{andrychowicz2020learning} achieved a breakthrough by using RL to demonstrate cube rotation with the Shadow Hand~\cite{ShadowRobotCompany2023ShadowSeries}. They applied domain randomization, which varied the simulation environment to help the robot perform better in real life. OpenAI~\cite{akkaya2019solving} also introduced automatic domain randomization (ADR), enabling the robot to solve a Rubik's cube. The use of GPU-based simulators like Isaac Gym has sped up the training process. Allshire et al.~\cite{allshire2022transferring} showed in-hand cube manipulation using the TriFinger \cite{Wuthrich2020TriFinger:Dexterity} robot. Similarly, Arunachalam et al. \cite{Arunachalam2022DexterousManipulation} and Qin et al.~\cite{Qin2022FromTeleoperation} utilized imitation learning to perform tasks on the Allegro Hand \cite{W.Robotics2023AllegroRd} like cube rotation and object flipping, using demonstrations captured with RGB and depth (RGB-D) cameras. Handa et al. \cite{handa2023dextreme} achieved cube reorientation using the Allegro Hand, with only 8 GPUs, outperforming earlier systems that relied on thousands of CPU cores. Their use of vectorized ADR enhanced the robot's ability to generalize by adding random variations that did not depend on physics. Yin et al. \cite{Yin2023RotatingTouch} used tactile feedback to enable the Allegro Hand rotate objects without visual input. Finally, Toshimitsu et al.~\cite{Toshimitsu2023GettingJoints} developed the Faive Hand, a biomimetic tendon-driven robotic hand. They demonstrated in-hand ball rotation after just one hour of training on a single NVIDIA A10G GPU. 
A learning-based alternative to reinforcement learning, behavioral cloning (BC), is rapidly emerging as data sets for human manipulation are becoming available through teleoperation solutions \cite{he2024omnih2o}. Most of these trained policies, as of now, are using simple parallel grippers such as Black et al. \cite{black2024pi_0} or Chi et. al \cite{chi2024universal}. With the growing industrial interest in humanoids, there is a trend of applying the same approaches to dexterous five-fingered robotic hands \cite{kareer2024egomimic}, but datasets of the same scale as simpler grippers \cite{o2024open} are not yet available. A proposed solution to this could be leveraging human data: Mandikal et al. have shown successful imitation learning of dexterous tasks from unstructured YouTube videos \cite{mandikal2022dexvip}, but as of now, this approach still needs hours of data for every task. 

Sampling-based MPC offers a promising alternative for dexterous manipulation where per-task retraining is not necessary. Bhardwaj et al. \cite{bhardwaj2022storm} successfully controlled high-degree-of-freedom robot arms for real-time tasks. Howell et al. \cite{howell2022predictive} achieved in-hand cube rotation in simulation using the MuJoCo physics engine on CPUs. Pezzato et al. \cite{pezzato2023sampling} used the GPU-parallelizable simulator Isaac Gym for non-prehensile manipulation with a physical robotic arm. While model-based control methods like MPC can solve complex tasks, they are computationally demanding and require precise modeling. Sampling-based methods do not depend on gradients, making them suitable for nonlinear and discontinuous dynamics. However, computational complexity limits its success in high-dimensional tasks. Current literature indicates that sampling-based MPC has not been applied to dexterous in-hand manipulation on physical robotic hands. 

\subsection{State of the art in objective function design via Large Language Models (LLMs)}

Objective functions are essential for directing robotic behavior and optimizing tasks. However, 92\% of researchers rely on manual trial-and-error methods for reward design, often yielding sub-optimal results \cite{Booth2023TheSpecifications}. Recent advancements in multimodal LLMs, and particularly VLMs, offer new solutions to these challenges. Ma et al. \cite{ma2023eureka} introduced Eureka, which leverages GPT-4's in-context learning to automate reward function design. This iterative process incorporates human language feedback to adapt rewards until satisfactory performance is achieved. Eureka successfully demonstrated pen spinning on a simulated Shadow Hand. Yu et al. \cite{yu2023language} used GPT-4 for objective function design and executed the task via sampling-based MPC. Their method facilitates human interaction by translating natural language commands into reward code, successfully achieving object grasping and drawer opening with a 7-degree-of-freedom arm and a jaw gripper. Liang et al. \cite{liang2024learning} further streamlined this process by embedding motion descriptions within a single prompt for reward code. This improved the teaching success rate for unseen tasks by 26.9\% and reduced human corrections, achieving successful object grasping using a physical 7-degrees-of-freedom (Dof) arm and a jaw gripper.

\subsection{Approach}

We use a tendon-driven robotic hand with bio-inspired rolling contact joints. This joint design enhances compliance but also introduces challenges in state estimation and modeling. We use sampling-based MPC with the MuJoCo physics engine to simulate the system dynamics. This approach simplifies task adaptation by allowing rapid modifications to task-specific objective functions. RL and BC methods require separate training for each task, while our approach can directly execute new objective functions. This allows an efficient integration into an evolutionary optimization loop without extensive retraining. We exploit the in-context learning capabilities of a VLM (GPT-4o) to perform evolutionary optimization of objective function weights with video feedback. We investigate dexterous manipulation tasks such as ball rolling, flipping, and catching. Our key contributions are:

\begin{itemize}
    \item We present the first successful demonstration of in-hand manipulation on a physical biomimetic tendon-driven robot hand using sampling-based MPC.
    \item We integrate a VLM into MPC to perform automated objective function tuning using video feedback, reducing manual objective function design.
    \item We demonstrate a successful adaptation of sampling-based MPC to real robots and dynamic tasks, showing the ability to roll, catch, and flip balls using a robotic hand.
\end{itemize}
Previous works explored the application of MPC or MPC-inspired algorithms to dexterous manipulation problems
\cite{howell2022predictive, jin2024task, morgan2020object}, but our work is the first to deploy sampling-based MPC on complex and compliant robotic hands.

%% file: Content/03-methodology.tex
\section{Methodology}

\begin{figure}[!b]
   \centering\includegraphics[width=1\columnwidth]{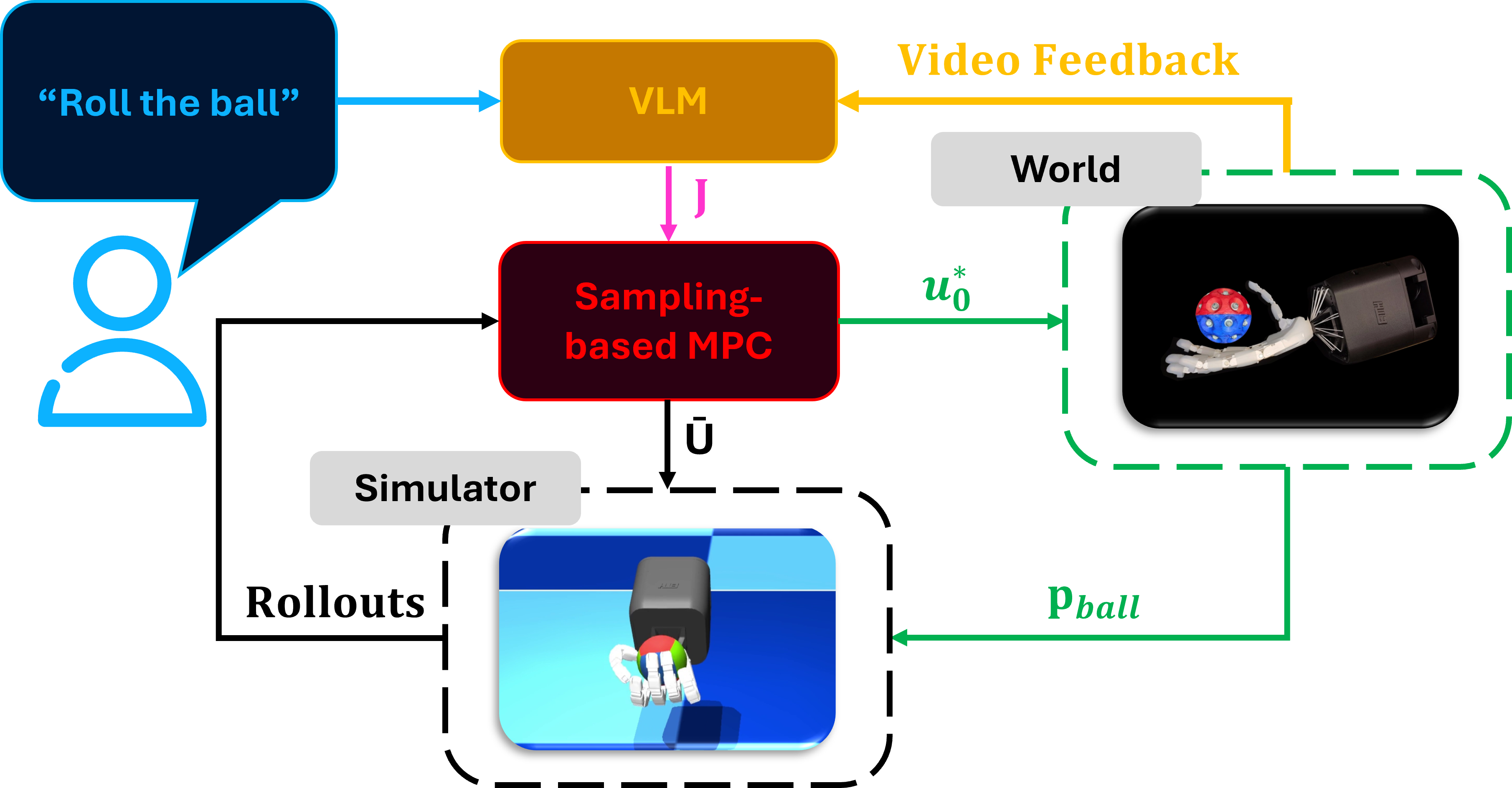}
   \caption{We propose the following pipeline to integrate VLMs with sampling-based MPC to control physical robot hands.}
   \label{fig:proposed-pipeline}
\end{figure}

 An overview of our proposed pipeline is shown in \Cref{fig:proposed-pipeline}. Our method uses the MuJoCo simulator as the dynamics model for sampling-based MPC \cite{howell2022predictive}. At each timestep, the ball position in the simulation is reset to the measured ball position $p_{ball}$, and randomized input sequences $\bar{U}$ are applied to parallel environments. The resulting rolled-out trajectories from the simulator are used to approximate the optimal control $u_{0}^{*}$ given the objective $J$. The optimal control $u_{0}^{*}$ (i.e., the desired joint angles) is used by the low-level controller of the robotic hand. The low-level controller of the robot hand calculates the necessary tendon lengths to achieve the desired joint angles, and commands the motors to the right angles to achieve them. The in-context learning capabilities of a multimodal LLM are used to perform evolutionary optimization of the objective function weights (i.e., parameters of $J$) using video feedback of the resulting behavior.

\subsection{Modelling and controlling the tendon-driven hand}

As described in \cite{Toshimitsu2023GettingJoints}, the Faive Hand has a biomimetic rolling contact joint design that mimics the motion of human fingers without fixed axes of rotation. In this work, we use the MuJoCo model of the Faive Hand provided by the authors to simulate its complex, tendon-driven actuation. Each rolling contact joint is represented by a pair of virtual hinge joints. The axes of these hinge joints are positioned to pass through the axis of the cylinder that constitutes each rolling contact surface. These hinge joints are constrained by tendons and fixed elements in the MJCF format to ensure coordinated motion. The tendon parameters, including stiffness and damping, enable the hand's rolling contact dynamics to be accurately simulated. The model also directly specifies the minimum and maximum joint angles for each joint, providing explicit joint angle constraints during simulation. The Faive Hand's joints are controlled by a low-level controller that translates desired joint angle commands into tendon length targets using a geometric model of the joint and tendon paths. An extended Kalman filter (EKF) uses tendon length measurements to estimate joint angles in real time, providing accurate proprioceptive feedback for low-level control. A detailed description of the low-level controller can be found in \cite{Toshimitsu2023GettingJoints}.

\subsection{Sampling-based MPC}

MPC is an optimization-based control approach that generates a sequence of control inputs by solving a receding-horizon optimization problem. It minimizes a cost function subject to system dynamics and constraints. Traditional gradient-based MPC relies on differentiable models and linear approximations, which struggle with the discontinuities and contact dynamics inherent in dexterous manipulation. In contrast, sampling-based MPC directly samples control trajectories, evaluating them in a physics simulator to determine the optimal action. This makes it well-suited for high-dimensional, contact-rich tasks where differentiability is impractical. We chose the MuJoCo physics engine to simulate the system dynamics due to its superior performance for real-time predictive control. While Isaac Gym excels in simulating simpler dynamics models thanks to GPU parallelization, our initial experiments showed that it struggles to meet simulation speed requirements for complex models such as the Faive Hand, limiting its effectiveness for real-time MPC applications.

Sampling-based MPC integrates optimal control principles with derivative-free optimization techniques to achieve effective trajectory planning and control in real-time scenarios. The approach involves selecting controls $u$ to minimize future costs or maximize returns in a dynamical system characterized by the state $x \in {\mathbb{R}}^n$, evolving according to:
\begin{equation}
x_{t+1} = f(x_t, u_t)
\end{equation}
where $ x_{t+1}$ is the updated state. The running cost is:
\begin{equation}
c(x_t, u_t)
\end{equation}
which can incorporate time-dependent factors through state integration. For a finite time horizon $T$, the objective function $J$ is given by:
\begin{equation}
J(x_{0:T}, u_{0:T}) = \sum_{t=0}^{T} c(x_t, u_t),
\end{equation}
where $ x_{0:T} $ and $ u_{0:T}$ denote sequences over discrete time.

Sampling-based MPC optimizes $ u_{0:T} $ as decision variables, while enforcing the dynamics through forward simulation. 
 In real-time operation, the current state $x_0$ is continuously estimated or measured, and the MPC optimizer computes 
 near-optimal controls $u_{0:T}$  starting from the currently estimated state $x_0$. 
 Instead of representing $u_{0:T}$ as $T+1$ independent control variables, we represent it using a spline function of order $O$, with spline parameters $\theta$ that need to be defined at $S$ distinct timepoints. This reduces the dimensionality of the search space for optimal controls from $T+1$ to $S$.   
 We use the Predictive Sampling method, which was introduced as an elementary baseline but proved to be surprisingly competitive~\parencite{howell2022predictive}. 

Predictive Sampling works as follows: A nominal sequence of actions, represented by spline parameters, is iteratively improved by random search. At each iteration, $K$ candidate splines are evaluated: The nominal itself and $K-1$ noisy samples from a Gaussian with the nominal as mean and a fixed standard deviation $\sigma$. In this work, we sample directly from the joint space of the 11 DoF Faive Hand. After sampling, the actions are clamped to the control limits by clamping the spline parameters $\theta$. Each candidate spline (i.e., controls) is rolled out in simulation starting from $x_0$. The total return, i.e., objective $J$, of each candidate is computed and the nominal spline is updated with the best candidate according to the objective. 

As in Howell et al. \cite{howell2022predictive}, we use the following form to describe the cost terms of our objective function for the different tasks:
\begin{equation}\label{eq:cost_term}
c(x, u) = \sum_{i=0}^{M-1} w_i \cdot n_i(r_i(x, u))
\end{equation}
This cost is a sum of $M$ terms, each comprising:
\begin{itemize}
    \item A weight $w \in {\mathbb{R}}^+$ determining the relative importance of this term.
    \item A norm $n(\cdot) : {\mathbb{R}}^d \to {\mathbb{R}}^+$, taking its minimum at $0_d$.
    \item The residual $r \in {\mathbb{R}}^d$ is a vector of elements that are “small” when the task is solved. 
\end{itemize}

\subsection{Evolutionary Adaptation with video feedback}\label{sec:vlm-adaptation}

We exploit the in-context learning capabilities of a VLM (GPT-4o) to perform evolutionary adaptation of objective function weights, $w_i$, with video feedback, without a human in the loop. The new objective function weights, as suggested by the VLM, are directly used by the sampling-based MPC, enabling real-time adaptation with each iteration taking only \qty{2}{min}. The adaptation process works as follows:  

\begin{enumerate}
    \item \textbf{Context Initialization:} The VLM is prompted with contextual information instructing it to act as a cost function engineer. It receives a structured Python template describing objective function weights, which serves as a baseline for its response. This step takes approximately \qty{15}{s}.
    \item \textbf{Task Definition:} A high-level description of the desired task is added by the user. For ball rolling we used the task description: ``Rotate the ball while ensuring it does not fall down" and for ball flipping we used the task description: ``Flip the ball while ensuring it does not fall down".
    \item \textbf{Objective Weight Generation:} In the first iteration, the VLM initializes the objective function weights based on the provided task definition.
    \item \textbf{Evaluation Strategy:} The VLM determines success criteria based on task descriptions and the generated weights. The VLM analyzes expected outcomes, key observations, and how different weights influence behavior. This step takes approximately \qty{15}{s}.
    \item \textbf{Execution \& Feedback Collection:} The sampling-based MPC controller runs with the new weights for \qty{30}{s}, and the resulting behavior is recorded as a video for \qty{10}{s}.
    \item \textbf{Reflection \& Iteration:} The recorded video, along with a reflection prompt, is fed back into the VLM. The model analyzes the results, compares them to previous iterations, and refines the cost weights, $w_i$, accordingly. This step takes approximately \qty{15}{s}. Steps 4–6 repeat until the robot successfully achieves the task. After each iteration the program sleeps for \qty{1}{min} to avoid running into token limitations.
\end{enumerate}

By leveraging this iterative adaptation strategy, the robot can autonomously learn new manipulation behaviors, refining its control strategy without predefined task-specific heuristics. This method enables rapid adaptation to different tasks while ensuring efficient learning cycles through systematic weight adjustments. The prompts for the context initialization, objective weights generation, and reflection are provided under \url{https://drive.google.com/file/d/1Xh5mX-uZsybAgeekGUFU6mbZt7G6VdK2}.

%% file: Content/04-experiments-results.tex
\section{Experiments and results}


In this section, we present our experimental results, supplemented by a video showcasing key simulations and real-world trials. To determine the weights of the objective functions for each task, we evaluate two alternative approaches. First, an Exhaustive Search method finds effective weight values by systematically testing weight combinations and recording performance metrics. The best-performing weights maximize a task-specific score, which, for ball rolling, includes average rotational velocity and ball drops per minute, while for ball flipping, it considers flip height, flips per minute, and ball drops per minute. 
For each task, exhaustive search tests 121 different weight combinations, where each combination is tested for ten minutes, resulting in a total of 20+ hours search for a good combination of weights.
Second, we employ Evolutionary Adaptation with video feedback (as described in Sec.~\ref{sec:vlm-adaptation}), leveraging a VLM to refine objective function weights for both tasks. Table~\ref{table:general-weights-all-tasks} lists the cost terms (Eq.~\ref{eq:cost_term}) used in the experiments, which are linearly combined to express different in-hand manipulation tasks, with the VLM autonomously assigning zero weights to irrelevant terms.

We used different parameters for Predictive Sampling depending on the task, as shown in Table~\ref{table:task-parameters}. For the ball flipping task, we increased $\sigma$ to allow for quicker reactions. For tasks involving only the robotic hand, a zero-order spline was sufficient, given the hand’s high responsiveness and the low-level controller used. However, for the ball-flipping task involving both the hand and the robot arm, we used a quadratic spline to achieve smoother arm trajectories.

\begin{table}[b!]
\centering
\caption{Predictive Sampling parameters for different manipulation tasks. The parameters include spline order ($O$), number of spline timepoints ($S$), number of candidate splines ($K$), horizon length ($T$), and standard deviation ($\sigma$).}
\label{table:task-parameters}
\begin{tabular}{lccccc}
\toprule
\textbf{Task} & $\bm{O}$ & $\bm{S}$ & $\bm{K}$ & $\bm{T}$ & $\bm{\sigma}$ \\ 
\midrule
Ball rolling & 0  & 5 & 10 & 25 & 0.1 \\ 
Ball flipping & 0 & 5 & 10 & 25 & 0.2 \\ 
Ball flipping with a robot arm & 2 & 8 & 10 & 25 & 0.1 \\ 
Ball catching & 0 & 5 & 10 & 25 & 0.1 \\ 
\bottomrule
\end{tabular}

\end{table}

\begin{table}[t!]
\centering
\caption{Different cost terms (Eq. 4) used in this paper. $\mathbf{w}$ is a weight vector, $w$ is a weight scalar, $\mathbf{p}$ is a position vector and $\mathbf{q}$ is the orientation quaternion.}
\label{table:general-weights-all-tasks}
\begin{tabular}{l l}
\toprule
\textbf{Name} & \textbf{Cost term} \\
\midrule
In Hand & $\|\mathbf{w}_{\text{InHand}} \cdot (\mathbf{p}_{\text{ball}} - \mathbf{p}_{\text{optimal}})\|_2^2$ \\ 
Ball Orientation & $ {w}_{\text{Orientation}} * \| \mathbf{q}_{\text{ball}} - \mathbf{q}_{\text{target}} \|_2^2$ \\ 
Ball Linear Velocity & $\|\mathbf{w}_{\text{LinVel}} \cdot \mathbf{v}_{\text{ball}}\|_2^2$ \\ 
Ball Height & $ {w}_{\text{Height}}* | h_{\text{ball}} - h_{\text{desired}} |$ \\ 
Flat Hand Configuration & $ {w}_{\text{FlatHand}}* \| \mathbf{q}_{\text{hand}} - \mathbf{q}_{\text{flat}} \|_2^2$ \\ 
Hold Ball Hand Configuration & ${w}_{\text{HoldBall}}*\| \mathbf{q}_{\text{hand}} - \mathbf{q}_{\text{hold}} \|_2^2$ \\ 
Faive Actuator & ${w}_{\text{FaiveActuator}}*\| \dot{\mathbf{q}}_{\text{hand}} \|_2^2$ \\ 
Panda Actuator & ${w}_{\text{PandaActuator}}*\| \dot{\mathbf{q}}_{\text{arm}} \|_2^2$ \\ 
\bottomrule
\end{tabular}
\end{table}

\begin{table*}[htbp]
\centering
\caption{Comparison of main results in simulation (optimal and evolutionary weights) and on the physical robot hand.}
\label{tab_overview-results}
\begin{tabular}{lccc}
\toprule
\textbf{Task} & \textbf{Sim. (Optimal Weights)} & \textbf{Sim. (Evol. Weights)} & \textbf{Real (Optimal Weights)} \\ 
\midrule
\textbf{Ball rolling} & \qty{1}{rad/s}, no drops & \qty{0.99}{rad/s}, 0.6 drops/min & \qty{0.35}{rad/s}, occasional drops \\
\textbf{Ball flipping} & 9.6 cm, 48 flips/min, 1 drop/min & 14 cm, 91 flips/min, 33 drops/min & Frequent drops \\
\textbf{Ball flipping (robot arm)} & 11.4 cm, 100 flips/min, no drops & Not investigated & Not feasible \\
\textbf{Ball catching} & Not investigated & Not investigated & 67\% success rate \\ 
\bottomrule
\end{tabular}
\end{table*}

\subsection{Simulation experiments}

We performed simulated experiments using the Faive Hand P0 with 11 DoFs, and a ball with a diameter of \qty{6.5}{cm}. The Faive Hand is fixed at an angle of 20 degrees relative to the horizontal plane for the rolling and flipping tasks. We further explored how fixing the Faive Hand on a robotic arm would change the flipping performance. In this case we fixed the Faive Hand on a Panda arm and, along with the hand joints, actuated joint number 4 and 6 of the arm. 

\begin{figure}[!b]
   \centering\includegraphics[width=1\columnwidth]{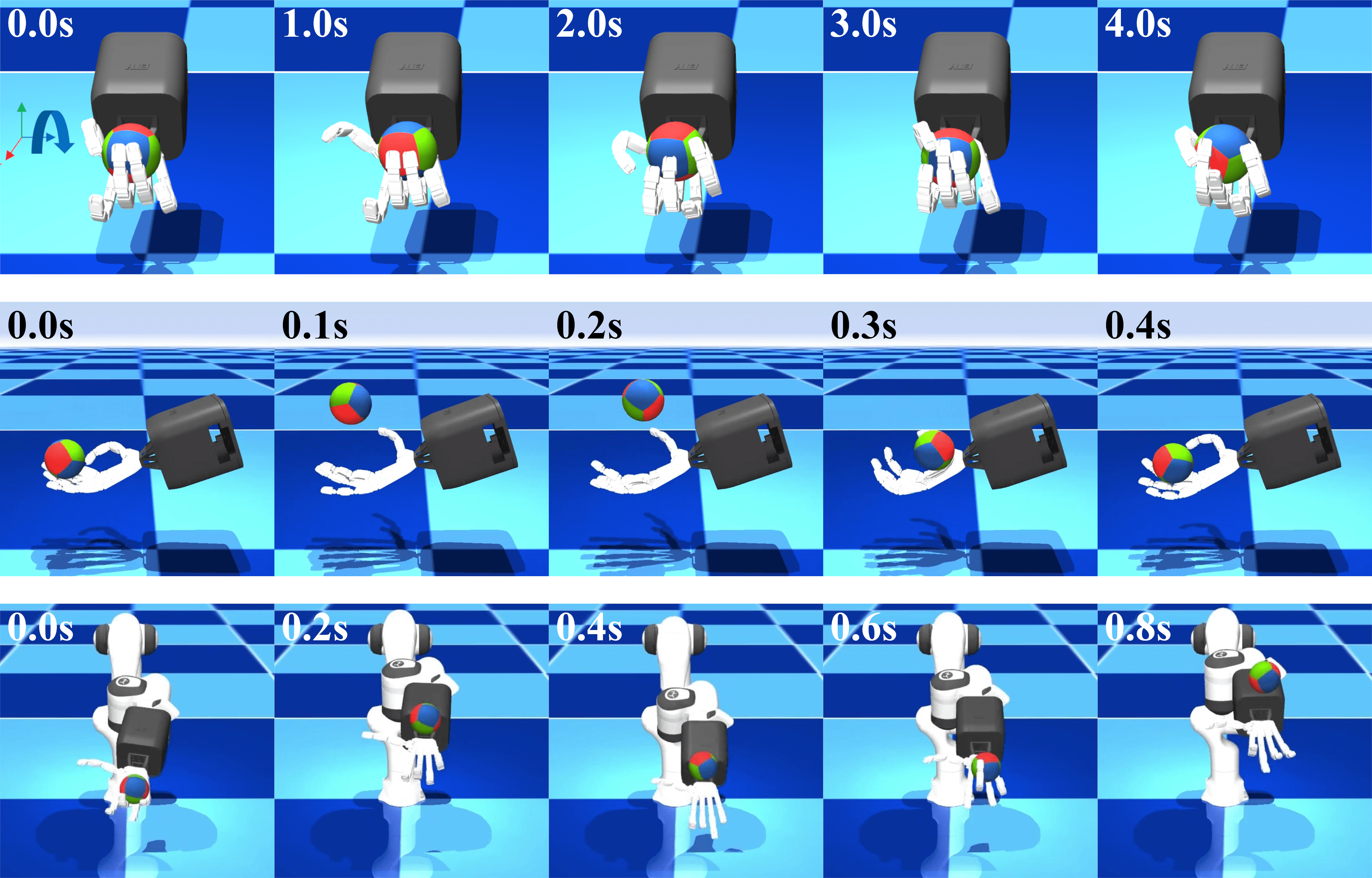}   \caption{We demonstrate ball rolling (top), ball flipping (middle) and flipping with a robot arm in simulation (bottom).}\label{fig:sim-experiments}
\end{figure}

\subsubsection{Ball rolling}
In this task, the goal is to roll a ball with a single Faive Hand. The desired rotational velocity is \qty{1}{rad/s}. The best objective function weights identified by the Exhaustive Search are: \( \mathbf{w}_{\text{InHand}} = [1000, 1000, 1000] \), \( w_{\text{Orientation}} = 20 \), \( \mathbf{w}_{\text{LinearVelocity}} = [10, 10, 10] \), \( w_{\text{Height}} = 0 \), \( w_{\text{FlatHand}} = 0 \), \( w_{\text{HoldBall}} = 10 \), and \( w_{\text{FaiveActuator}} = 0.1 \).

Using these weights, ball rolling was achieved around all axes. \Cref{fig:sim-experiments} shows rolling around the negative y-axis. Ball rolling was executed smoothly and consistently. The ball remained securely in the hand throughout the simulations and never fell down.  We achieved stable rolling with average rotational velocities: \( \omega_x = 0.96 \, \text{rad/s} \), \( -\omega_x = 0.97 \, \text{rad/s} \), \( \omega_y = 0.96 \, \text{rad/s} \), \( -\omega_y = 0.96 \, \text{rad/s} \), \( \omega_z = 1.01 \, \text{rad/s} \), and \( -\omega_z = 0.99 \, \text{rad/s} \), without the ball falling off. Given the hardware limitations, \textit{e.g.}, the lack of abduction and adduction, as well as inaccurate thumb placement, it is surprising that rolling around the x and z-axis work as effectively as around the y-axis.

The objective function weights after four Evolutionary Adaptation cycles (taking 8 minutes) are: \( \mathbf{w}_{\text{InHand}} = [50, 50, 50] \), \( w_{\text{Orientation}} = 900 \), \( \mathbf{w}_{\text{LinearVelocity}} = [0, 0, 0] \), \( w_{\text{Height}} = 0 \), \( w_{\text{FlatHand}} = 0 \), \(w_{\text{HoldBall}} = 300 \), and \( w_{\text{FaiveActuator}} = 0.1 \).
The achieved performance is a rotational velocity of \( -\omega_x = 0.97 \, \text{rad/s} \) and the ball drops 0.6 times per minute. This suggests that while the VLM was able to find a solution for rolling the ball, it did not identify the most stable configuration.

\subsubsection{Ball flipping}

In this task, our goal is to flip the ball \qty{15}{cm} high and catch it with a single Faive Hand. A dynamic flipping motion should be observed where the ball is constantly thrown up and caught again. The best weights identified by the Exhaustive Search are: \( \mathbf{w}_{\text{InHand}} = [500, 500, 0] \), \( w_{\text{Orientation}} = 0 \), \( \mathbf{w}_{\text{LinearVelocity}} = [40, 80, -120] \), \( w_{\text{Height}} = 140 \), \( w_{\text{FlatHand}} = 20 \), \( w_{\text{HoldBall}} = 0 \), and \( w_{\text{FaiveActuator}} = 0.1 \).

 With these weights, we ran the experiments for ten minutes and recorded the flip height, flip rate, drop rate. We achieved successful ball flipping in simulation with an average flip height of \qty{9.6}{cm}. This is considerably lower than the desired flip height of \qty{15}{cm}. We achieved a dynamic flipping motion as shown in \Cref{fig:sim-experiments} but sometimes the ball got stopped in the palm of the hand and the continuous flipping motion was interrupted. This is the reason why the flip rate was only 48 flips per minute.  The flipping behavior is quite stable and the ball drops only once per minute. The flip rate was measured as the number of successful flips per minute in which the ball completed an upward trajectory and returned to the palm. If the ball fell to the ground, it was reinitialized in the palm, and the count was not resumed until the flip sequence was restored.
 
The objective function weights after four Evolutionary Adaptation cycles (taking 8 minutes) are: \( \mathbf{w}_{\text{InHand}} = [800, 800, 0] \), \( w_{\text{Orientation}} = 0 \),
\( \mathbf{w}_{\text{LinearVelocity}} = [0, 0, -900] \), \( w_{\text{Height}} = 1000 \), \( w_{\text{FlatHand}} = 900 \), \( w_{\text{HoldBall}} = 0 \), and \( w_{\text{FaiveActuator}} = 0.1 \). The resulting average flip height is \qty{14.0}{cm} with a flip rate of 91 flips per minute. The ball drops on average 33 times per minute. The flip height is significantly higher than what the Exhaustive Search weights achieved (only 9.6 cm). Compared to our the Exhaustive Search weights, the ball drops more frequently. We can again observe the tendency to prioritize achieving a better performance over not dropping the ball.

\subsubsection{Ball flipping with a robot arm}
Given the ball height achieved in the previous set of experiments, we also investigated whether giving more degrees of freedom to the MPC, by incorporating robot arm joints, would create a significant change in performance.
Hence, in this task, our goal is to throw the ball to a target height of 8-\qty{12}{cm} and catch it with a Faive Hand mounted on a Panda arm. A dynamic flipping motion should be observed where the ball is constantly thrown up and caught again. This task uses a second-order spline with eight spline points to achieve smoother actions compared to other tasks (\Cref{table:general-weights-all-tasks}). The best weights identified by the Exhaustive Search are: \( \mathbf{w}_{\text{InHand}} = [750, 750, 0] \), \( w_{\text{Orientation}} = 0 \),
\(\mathbf{w}_{\text{LinearVelocity}} = [0, 0, -5] \), \( w_{\text{Height}} = 10 \), \( w_{\text{FlatHand}} = 20 \), \( w_{\text{HoldBall}} = 0 \), \( w_{\text{FaiveActuator}} = 0.1 \), and \( w_{\text{PandaActuator}} = 20 \). We ran the experiments with these weights for ten minutes and recorded the flip height, flip rate, drop rate. We achieved successful ball flipping with an average flip height of \qty{11.4}{cm}. The ball flipping resulted in a dynamic flipping motion as shown in \Cref{fig:sim-experiments} and the ball never fell out of the hand. Flipping the ball with the robotic arm resulted in 100 flips per minute in a continuous flipping motion. Compared to flipping with the robotic hand alone, the system with the robotic arm achieves a higher flip rate and never drops the ball.


 \subsection{Demonstrations on the physical robot hand}

Next, we investigated the performance of MPC on the physical robot system. Here, since our main goal is to evaluate the MPC performance on physical dexterous manipulation tasks, and since the high number of executions required for Exhaustive Search and Evolutionary Adaptation are difficult to realize on the physical robot, we used a set of weights already optimized from simulation. Since Exhaustive Search weights resulted in fewer drops, we used those weights for the demonstrations on the physical hand.

For the demonstrations, we used the Faive Hand P0 with 11 DoFs attached to a static mount and a ball with a diameter of \qty{6.5}{cm}. The Faive Hand P0 has a total length of \qty{20.1}{cm}, a palm width of \qty{8.1}{cm}, and a finger length of \qty{12.2}{cm} to \qty{14.0}{cm}. The Faive Hand is positioned at an angle of \qty{20}{^{\circ}} relative to the horizontal plane. It is equipped with silicone padding around the fingers and on the fingertips. The Faive Hand P0 can only indirectly estimate joint angles from tendon lengths, which results in limited proprioceptive accuracy. Therefore, we do not provide joint angle feedback to the MPC controller. Instead, the system leverages the hand's inherent compliance, which is essential for performing dexterous manipulation tasks. This compliance allows the fingers to adapt to object geometries and maintain stable contact without requiring precise estimation of the joint state. Research has shown that compliance naturally handles disturbances and positional uncertainties, reducing the need for detailed contact modeling \cite{Bhatt__2021}. It also allows the hand to respond naturally to external forces and interactions, supporting robust, human-like grasping behaviors \cite{Santina2018TowardSynergies}. However, for tasks such as catching and flipping, where the ball leaves the hand, it is crucial to track the ball’s position with high frequency and precision to enable accurate control. To support these tasks, our ball is equipped with 28 evenly spaced reflective markers. The Qualisys Motion Capture (MoCap) System, equipped with eight cameras, is used to track reflective markers at a frequency of 100 Hz. An additional camera captures RGB video. The Qualisys MoCap System runs on an Intel NUC with an i7-10710U CPU. We use the MuJoCo physics engine to simulate the system dynamics. The simulation runs on a high-performance computer with Ubuntu 20.04 and an i9-10900K CPU with 20 threads.

The ball's position is updated in the simulation to match the measured state. To do this, we use random sample consensus (RANSAC) on the ball’s 28 reflective markers, allowing robust detection of the ball’s center, even in the presence of occlusions. We update the ball position in the simulation to the measured ball position at 30 Hz. The Predictive Sampling method simulates the task from the initialized point and predicts the best action according to the objective function. We then send the best action to the low-level joint controller of the hand at 30 Hz. The positional accuracy of our ball center detection system is within 5 mm, and the standard deviation of the ball center noise is less than 1 mm. We also observed that the detection remains accurate even when the ball is moving at high speed through the air. Due to the high noise and insufficient accuracy of orientation measurements, we do not perform orientation estimation. Instead, in each optimization step, the orientation of the ball in the simulation is kept fixed, and the target orientation is set to $60^\circ$ in front of this fixed orientation in the desired direction of rotation.

Sampling-based MPC does not strive to find the optimal solution in a single step but instead aims to make incremental progress towards the goal. This approach is well-suited for handling uncertainties and state estimation errors in real-time. It is often more important to replan based on updated state measurements after each iteration rather than attempting to converge to an outdated solution. The tendon-driven actuation, combined with this iterative approach, ensures that the system remains stable and compliant, even in the presence of state estimation uncertainties.

\subsubsection{Ball rolling}

\begin{figure}[!b]
   \centering\includegraphics[width=1\columnwidth]{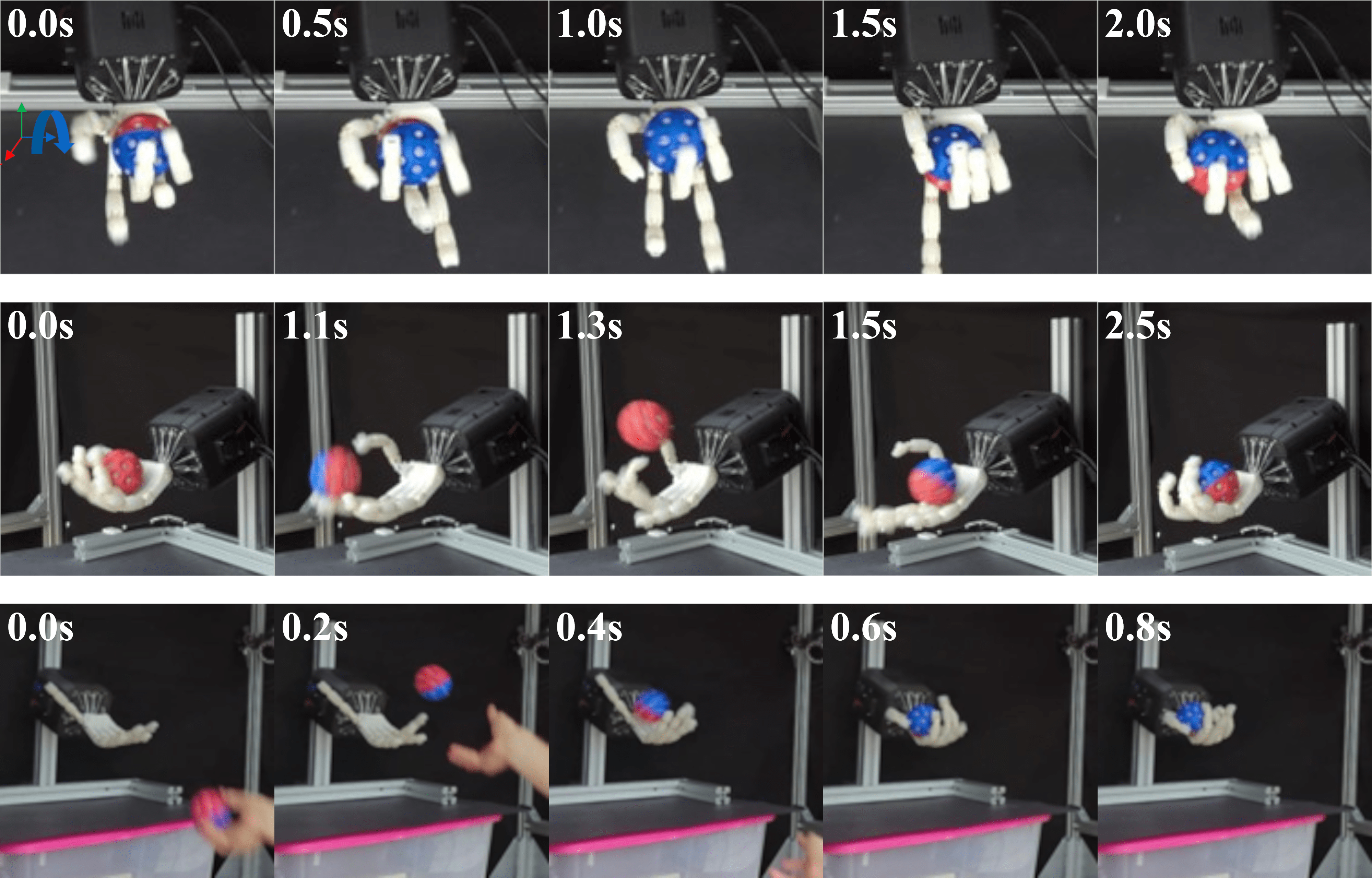}
   \caption{We demonstrate successful ball rolling, ball flipping and ball catching on a physical robot hand. Please also see our video: \url{https://youtu.be/6ivbd_jijHA}.}
   \label{fig:real-experiments}
\end{figure}

We demonstrate the feasibility of ball rolling around the negative y-axis with the Exhaustive Search weights. The rotational velocity was inconsistent and the ball occasionally fell out of the hand. A \qty{180}{^{\circ}} rotation was completed in \qty{2}{s}, as illustrated in \Cref{fig:real-experiments}. The rotation speed varied significantly, ranging from \qty{0.2}{rad/s} to \qty{1.6}{rad/s}, with an average speed of \qty{0.35}{rad/s} for a full rotation. We observe that the finger movements are very fast, but not very smooth. This can potentially be improved through sampling of larger number of trajectories (i.e., a higher $K$), though it would then also mean a lower MPC frequency due to the increased computational expense.  Rotation around the x-axis and z-axis could not be achieved on the physical robot hand. This was expected because such rotations are more difficult to perform without abduction and adduction capabilities, and results on the physical robot are sensitive to differences between simulation and reality.

We compare our ball-rolling results with those reported by Toshimitsu et al. \cite{Toshimitsu2023GettingJoints}. Their approach used RL with domain randomization to train a ball-rolling policy on an identical robot hand. They used joint angle estimations provided by an EKF as proprioceptive feedback for the RL policy, without incorporating the ball position into the policy. Using domain randomization, they successfully achieved smooth rotations around the y-axis in both directions. Their approach maintained a consistent rotational velocity of \qty{1.5}{rad/s}, which is significantly faster than our reported average rotational velocity of \qty{0.35}{rad/s}. The ball rotation motion generated by their RL policy was smoother than we observed with sampling-based MPC. However, a limitation of their method is the lack of ball position feedback, which simplified the policy by consistently repeating the same motion without considering the position of the ball. Incorporating ball position feedback into the policy would make their method more robust to external disturbances. It is also important to note that their RL policy would require an hour of training to adapt to changes in the task, whereas our approach adapts to a new task in only eight minutes.

\subsubsection{Ball flipping}
We show ball flips with a single Faive Hand using the Exhaustive Search weights, as shown in \Cref{fig:real-experiments}. With a target height of \qty{15}{cm}, heights greater than \qty{10}{cm} were often reached. During the demonstration, the flipping motion was inconsistent, resulting in frequent instances of the ball falling to either side of the palm. This task is particularly challenging due to its dynamic nature, which requires precise timing and coordination. Even minor deviations in the position of the ball or fingers in the real system compared to the simulation, caused by latency, sensor noise or inaccurate actuation, often result in failed flips. The exact timing of the flip relative to the ball's position is critical to successful execution, making the task extremely sensitive to small errors. Additionally, limitations of the robotic hand increase the difficulty and robustness of this task. First, the hand lacks wrist mobility, a critical feature that allows humans to perform complex manipulations, such as juggling, where dynamic wrist movements adjust hand orientation. In addition, the lack of abduction and adduction capabilities hinders the fingers' ability to accurately position themselves under the ball's center of gravity. Inaccuracies in thumb positioning further hinder the robot's ability to prevent the ball from rolling to the side.

\subsubsection{Ball catching}

 The ball catching task was performed directly on the physical robot. In this demonstration, the goal is to catch a ball using a five-fingered robotic hand, with its base attached to a static mounting bracket. A human operator throws the ball toward the center of the hand. The goal of the task for the robotic hand is to catch and maintain a hold of the ball. For this task, we directly adapt the weights for the use in the real environment and do not conduct tests in simulation. We used the weights (only reporting the non-zero weights): \( w_{\text{inHand}} = 1000 \), \( w_{\text{FaiveActuator}} = 0.1 \), and \( w_{\text{HoldBall}} = 100 \). We demonstrate the feasibility of catching a ball with a robot hand as seen in \Cref{fig:real-experiments}. We tried 36 throws that landed in the palm of the hand and managed to catch 24. This corresponds to a success rate of 67\,\%. This task is highly dependent on the trajectory and force of the ball thrown by the human tester, which caused variability in throws.

\subsection{Limitations}

During this work, we encountered several challenges related to computational efficiency, the complexity of the Faive Hand model, and hardware limitations of the robotic hand. The computational demands of sampling-based MPC influenced the smoothness of trajectory generation, particularly in tasks requiring rapid manipulation. The complexity of the Faive Hand model limited the use of GPU parallelization in Isaac Gym due to insufficient simulation steps per second for real-time manipulation. Hardware issues such as inaccurate proprioception, thumb positioning, and lack of abduction/adduction capabilities further limited performance. In addition, VLM's token-per-minute constraints reduced video frame rates, which impacted task recognition. Still, we were able to demonstrate some of the first examples of MPC-based control of a variety of dexterous manipulation tasks on a physical anthropomorphic robot hand.

%% file: Content/05-conclusion.tex
\section{Conclusion}

We presented a framework that uses sampling-based MPC to perform dexterous in-hand ball manipulation tasks. We demonstrate tasks such as ball rolling, ball flipping, and ball catching using a physical robotic hand and ball position feedback. By combining sampling-based MPC with VLMs and video feedback, we enabled rapid autonomous iteration and improvement of manipulation behaviors. Using our method, each iteration is completed in less than two minutes, allowing for rapid iteration cycles. After only a few iterations of evolutionary optimization with video feedback, the robot has successfully learned how to perform in-hand ball rotation and ball flipping. Despite these advances, challenges remain, including optimizing sampling efficiency in high-dimensional action spaces and improving proprioceptive accuracy. Future work will focus on real-time domain randomization, refine the action space for more efficient sampling, address hardware constraints, and implement direct evolutionary adaptation on the physical robotic hand.